\documentclass[11pt,a4paper]{article}
\usepackage{authblk}
\usepackage{graphicx}
\usepackage{naaclhlt2018}
\usepackage{times}
\usepackage{latexsym}
\usepackage{amsmath}
\usepackage{amsfonts}
\usepackage{mathtools}
\usepackage{microtype}
\usepackage[small]{caption}
\usepackage{subcaption}
\usepackage{float}
\usepackage[safe]{tipa}
\usepackage{booktabs}
\usepackage{xfrac}
\usepackage{appendix}
\usepackage{adjustbox}

\usepackage{algorithm}
\usepackage[noend]{algpseudocode}

\usepackage[disable]{todonotes}
\newcommand{\note}[4][]{\todo[author=#2,color=#3,size=\scriptsize,fancyline,caption={},#1]{#4}}
\newcommand{\jason}[2][]{\note[#1]{jason}{green!40}{#2}}
\newcommand{\Jason}[2][]{\jason[inline,#1]{#2}}
\newcommand{\ryan}[2][]{\note[#1]{ryan}{violet!40}{#2}}
\newcommand{\Ryan}[2][]{\ryan[inline,#1]{#2}}
\newcommand{\cc}[2][]{\note[#1]{chucheng}{brown!40}{#2}}

\newlength{\extramargin}
\makeatletter\if@todonotes@disabled\else
\usepackage{calc}
\usepackage[paperwidth=\paperwidth+\extramargin*2,marginparwidth=\marginparwidth+\extramargin,width=\textwidth,height=\textheight]{geometry}
\fi\makeatother

\renewcommand{\algorithmicindent}{9pt}
\algnewcommand{\LineComment}[1]{\State \(\triangleright\) {\small \it #1}}
\algnewcommand{\InlineComment}[1]{\hfill \(\triangleright\) {\small \it #1}}
\algrenewcommand\algorithmicindent{1.0em}%

\usepackage{authblk}
\usepackage{url}
\usepackage{tikz}
\usetikzlibrary{bayesnet}

\definecolor{darkgrey}{rgb}{0.2,0.2,0.2}
\definecolor{grey}{rgb}{0.9,0.9,0.9}
\definecolor{darkblue}{rgb}{0.0,0.0,0.5}
\definecolor{darkred}{rgb}{0.5,0.0,0.0}
\definecolor{darkorange}{rgb}{1.0,0.55,0.0}
\definecolor{darkgreen}{rgb}{0.0,0.6,0.0}
\definecolor{darkyellow}{rgb}{1.0,0.65,0.0}
\definecolor{darkorange}{rgb}{1.0,0.65,0.0}
\definecolor{darkergreen}{rgb}{0.0,0.4,0.0}
\definecolor{lightblue}{rgb}{0.8,0.8,1.0}
\definecolor{lightgreen}{rgb}{0.8,1.0,0.8}
\definecolor{lightred}{rgb}{1.0,0.8,0.8}
\definecolor{lightyellow}{rgb}{1.0,1.0,0.8}
\definecolor{lightorange}{rgb}{1.0,0.9,0.8}
\definecolor{lightgrey}{rgb}{0.96,0.97,0.98}
\definecolor{brilliantlavender}{rgb}{0.96, 0.73, 1.0}

\definecolor{ryanred}{rgb}{0.64, 0.0, 0.0}
\definecolor{ryanblue}{rgb}{0.13, 0.0, 0.58}
\definecolor{ryangreen}{rgb}{0.12, 0.59, 0.0}
\definecolor{ryanpurple}{rgb}{0.65, 0.0, 0.57}

\newcommand*{\numberingBlue}[1]{%
  \protect\tikz[baseline={([yshift=-1.5pt]n.base)}]%
  \protect\node[fill=blue!25,shape=circle,inner sep=1pt,draw](n){\tiny #1};}

\newcommand*{\numberingRed}[1]{%
  \protect\tikz[baseline={([yshift=-1.5pt]n.base)}]%
  \protect\node[fill=red!15 ,shape=circle,inner sep=1pt,draw](n){\tiny #1};}

\newcommand*{\numberingGreen}[1]{%
  \protect\tikz[baseline={([yshift=-1.5pt]n.base)}]%
  \protect\node[fill=darkgreen!25,shape=circle,inner sep=1pt,draw](n){\tiny #1};}

\newcommand*{\numberingYellow}[1]{%
  \protect\tikz[baseline={([yshift=-1.5pt]n.base)}]%
  \protect\node[fill=yellow!25,shape=circle,inner sep=1pt,draw](n){\tiny #1};}

\newcommand*{\numberingBlueB}[1]{%
  \protect\tikz[baseline={([yshift=-1.5pt]n.base)}]%
  \protect\node[fill=blue!25,shape=circle,inner sep=1pt,draw](n){\small #1};}

\newcommand*{\numberingRedB}[1]{%
  \protect\tikz[baseline={([yshift=-1.5pt]n.base)}]%
  \protect\node[fill=red!15 ,shape=circle,inner sep=1pt,draw](n){\small #1};}

\newcommand*{\numberingGreenB}[1]{%
  \protect\tikz[baseline={([yshift=-1.5pt]n.base)}]%
  \protect\node[fill=darkgreen!25,shape=circle,inner sep=1pt,draw](n){\small #1};}

\newcommand*{\numberingYellowB}[1]{%
  \protect\tikz[baseline={([yshift=-1.5pt]n.base)}]%
  \protect\node[fill=yellow!25,shape=circle,inner sep=1pt,draw](n){\small #1};}

\usepackage{cleveref}

\crefname{section}{ }{\S\S}
\Crefname{section}{\S}{\S\S}
\crefname{table}{Tab.}{Tables}
\crefname{figure}{Fig.}{Figs.}
\crefname{algorithm}{Alg.}{Algs.}
\crefname{equation}{eq.}{equations}
\crefname{appendix}{App.}{Appendices}
\crefformat{section}{\S#2#1#3}

\newcommand{\bigV}{{\cal V}}
\newcommand{\bigVbar}{\bar{\bigV}}

\newcommand{\Vbar}{\bar{V}}
\newcommand{\vbar}{\bar{v}}
\renewcommand{\l}{^\ell}

\aclfinalcopy

\usepackage{xcolor}
\makeatletter
\def\mathcolor#1#{\@mathcolor{#1}}
\def\@mathcolor#1#2#3{%
  \protect\leavevmode
  \begingroup
    \color#1{#2}#3%
  \endgroup
}
\makeatother

\usepackage{pifont}
\newcommand{\xmark}{\ding{55}}%

\definecolor{mylavender}{HTML}{BD71E1}
\definecolor{darkpurple}{HTML}{531B93}

\newcommand{\vtheta}{{\boldsymbol \theta}}

\newcommand{\vmu}{{\boldsymbol \mu}}

\newcommand{\vv}{{\mathbf{v}}}
\newcommand{\xx}{{\mathbf{x}}}
\newcommand{\vvtilde}{{\tilde{\mathbf{v}}}}
\newcommand{\va}{{\mathbf{a}}}
\newcommand{\NN}{\nu_{\vtheta}}

\title{A Deep Generative Model of Vowel Formant Typology}

\author{{\bf Ryan Cotterell} \and {\bf Jason Eisner} \\ Department of Computer Science \\ Johns Hopkins University, Baltimore MD, 21218 \\ \texttt{\{ryan.cotterell,eisner\}@jhu.edu}}

\begin{document}
\maketitle
\begin{abstract}
  What makes some types of languages more probable than others?
  For instance, we know that almost all spoken languages contain the
  vowel phoneme /i/; why should that be? The field of linguistic typology
  seeks to answer these questions and, thereby, divine the
  mechanisms that underlie human language. In our work,
  we tackle the problem of vowel system typology, i.e.,
  we propose a generative probability model of which vowels a language contains.
  In contrast to previous work, we work directly with the acoustic
  information---the first two formant values---rather than modeling discrete sets of phonemic symbols (IPA). We develop a novel generative
  probability model and report results based on a corpus of 233 languages.
\end{abstract}

\section{Introduction}
Human languages are far from arbitrary; cross-linguistically, they
exhibit surprising similarity in many respects and many properties appear to be universally
true. The field of linguistic typology seeks to investigate, describe
and quantify the axes along which languages vary. One facet of
language that has been the subject of heavy investigation is the
nature of vowel inventories, i.e., which vowels a language
contains. It is a cross-linguistic universal that all spoken languages have
vowels \cite{gordon}, and the underlying principles guiding vowel selection are
understood: vowels must be both easily recognizable
and well-dispersed \cite{schwartz2005dispersion}. In this
work, we offer a more formal treatment of the subject, deriving a generative probability model of vowel inventory typology.  Our work builds on
\cite{cotterell-eisner:2017:ACL2017} by investigating not just
discrete IPA inventories but the cross-linguistic variation
in acoustic formants.

The philosophy behind our approach is that linguistic typology
should be treated probabilistically and its goal should be the construction of a universal prior
over potential languages.
A probabilistic approach does not rule out linguistic systems
completely (as long as one's theoretical formalism can describe them
at all), but it can position phenomena on a scale from very common to
very improbable.  Probabilistic modeling also provides a discipline
for drawing conclusions from sparse data.  While we know of over 7000
human languages, we have some sort of linguistic analysis for only
2300 of them \cite{wals-s1}, and the dataset used in this paper
\cite{becker2010acoustic} provides simple vowel data for fewer than
250 languages.

Formants are the resonant frequencies of the human vocal tract during
the production of speech sounds.
We propose a
Bayesian generative model of vowel inventories, where each language's inventory
is a finite subset of acoustic vowels represented as points $(F_1,F_2)
\in \mathbb{R}^2$.
We deploy tools from the neural-network and
point-process literatures and experiment on a dataset
with 233 distinct languages.  We show that
our most complicated model outperforms simpler
models.

\section{Acoustic Phonetics and Formants}\label{sec:phonetics}
Much of human communication takes place through speech: one conversant
emits a sound wave to be comprehended by a second. In this work, we
consider the nature of the portions of such sound waves that correspond
to vowels. We briefly review the relevant bits of acoustic
phonetics so as to give an overview of the data we are actually
modeling and develop our notation.

\begin{figure}
  \centering
  \includegraphics[width=1.\columnwidth]{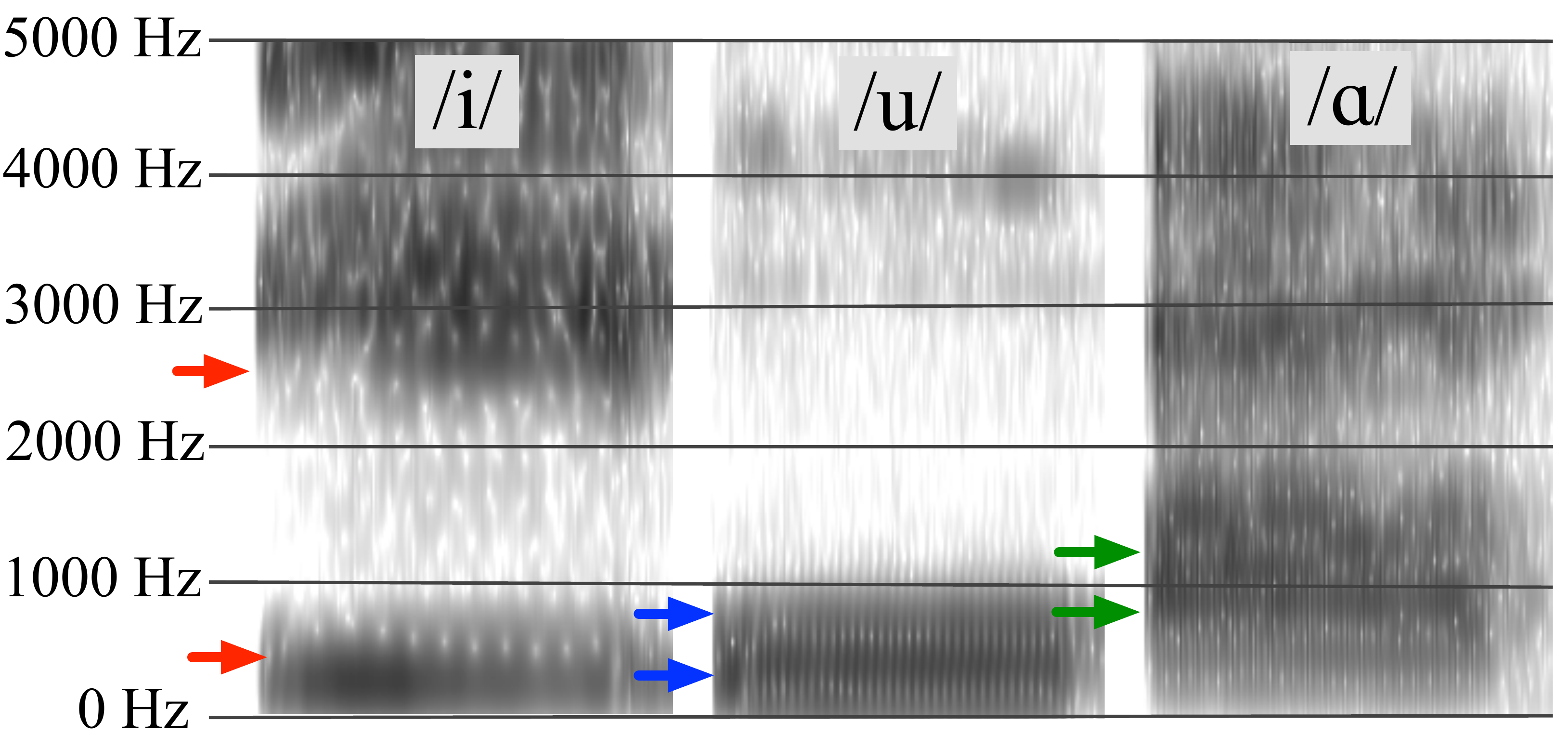}
  \caption{Example spectrogram of the three English vowels: /i/, /u/ and /\textipa{A}/. The $x$-axis is time and $y$-axis is frequency. The first two formants $F_1$ and $F_2$
    are marked in with arrows for each vowel. The figure was made with Praat \cite{boersma2002praat}.}
  \label{fig:spectrogram}
\end{figure}

\paragraph{The anatomy of a sound wave.}
The sound wave that carries spoken language is a function from time to
amplitude, describing sound pressure variation in the air. To distinguish
vowels, it is helpful to transform this function into a {\bf spectrogram}
(\cref{fig:spectrogram}) by using a short-time Fourier transform \cite[Chapter 1]{deng2003speech}
to decompose each short interval of the wave function into a weighted sum of sinusoidal waves of different
frequencies (measured in Hz).  At each interval, the variable darkness of the spectrogram indicates the weights
of the different frequencies. In phonetic
analysis, a common quantity to consider is a {\bf formant}---a local maximum
of the (smoothed) frequency spectrum. The fundamental frequency $F_0$ determines
the pitch of the sound. The formants $F_1$ and $F_2$ determine the quality of the vowel.

\paragraph{Two is all you need (and what we left out).}
In terms of vowel recognition, it is widely speculated that humans
rely almost exclusively on the first two formants of the sound wave
\cite[Chapter 5]{ladefoged2012vowels}. The two-formant assumption breaks down in edge cases: e.g., the
third formant $F_3$ helps to distinguish the roundness of the vowel
\cite[Chapter 5]{ladefoged2012vowels}.  Other non-formant
features may also play a role. For example, in tonal languages, the same
vowel may be realized with different tones (which are signaled using $F_0$):
Mandarin Chinese makes a distinction between m\textipa{\v{a}} ({\em horse}) and m{\'a} ({\em hemp}) without
modifying the quality of the vowel /a/. Other features, such as creaky
voice, can play a role in distinguishing phonemes. \cc{if you want to cut space, maybe just say there are some non-formant features such as tones etc. no need to go into details.} We do not
explicitly model any of these aspects of vowel space, limiting ourselves to $(F_1, F_2)$
as in previous work \cite{liljencrants1972numerical}. However, it would be easy to extend
all the models we will propose here to incorporate such information, given
appropriate datasets.

\section{The Phonology of Vowel
Systems}\label{sec:phonology}
The vowel inventories
of the world's languages display clear structure
and appear to obey several underlying principles. The most prevalent
of these principles are \textbf{focalization} and \textbf{dispersion}.
\paragraph{Focalization.}
The notion of focalization grew out of quantal vowel theory
\cite{stevens1989quanta}.  Quantal vowels are those that are phonetically
``better'' than others.  They tend to display certain properties,
e.g., the formants tend to be closer together
\cite{stevens1987relational}. Cross-linguistically, quantal vowels are the most
frequently attested vowels, e.g., the cross-linguistically common
vowel /i/ is considered quantal, but
less common /y/ is not.

\paragraph{Dispersion.}
The second core principle of vowel system organization is known as
dispersion. As the name would imply, the principle states that
the vowels in ``good'' vowel systems tend to be spread out. The motivation
for such a principle is clear---a well-dispersed set of vowels
reduces a listener's potential confusion over which vowel is being
pronounced. See \newcite{schwartz1997dispersion}
for a review of dispersion in vowel system typology and its interaction
with focalization, which has led to the joint dispersion-focalization theory.

\paragraph{Notation.}
We will denote the universal set of international phonetic alphabet (IPA) symbols as $\bigV$.  The observed vowel inventory for language $\ell$ has size $n\l$ and is denoted $V\l = \{(v\l_1,\vv\l_1), \ldots, (v\l_{n\l},\vv\l_{n\l})\} \subseteq \bigV \times \mathbb{R}^d$, where for each $k \in [1,n\l]$, $v\l_k \in \bigV$ is an IPA symbol assigned by a linguist and $\vv\l_k \in \mathbb{R}^d$ is a vector of $d$ measurable phonetic quantities. In short, the IPA symbol $v\l_k$ was assigned as a label for a phoneme with pronunciation $\vv\l_k$.  The ordering of the elements within $V\l$ is arbitrary.

\paragraph{Goals.}  This framework recognizes that the same IPA symbol $v$ (such as /u/) may represent a slightly different sound $\vv$ in one language than in another, although they are  transcribed identically.  We are specifically interested in how the vowels in a language influence one another's fine-grained pronunciation in $\mathbb{R}^d$.
 In general, there is no reason to suspect that
 speakers of two languages, whose phonological systems contain the same
 IPA symbol, should produce that vowel with identical formants.

\paragraph{Data.} For the remainder of the paper, we will take $d=2$ so that each $\vv = (F_1,F_2) \in \mathbb{R}^2$, the vector consisting of the first two formant values, as compiled from the field literature by \newcite{becker2006predicting}.  This dataset provides inventories $V\l$ in the form above.  Thus, we do not consider further variation of the vowel pronunciation that may occur within the language (between speakers, between tokens of the vowel, or between earlier and later intervals within a token).

\section{Phonemes versus Phones}

Previous work \cite{cotterell-eisner:2017:ACL2017} has placed a
distribution over discrete phonemes, ignoring the variation
across languages in the \emph{pronunciation} of each phoneme.  In this
paper, we crack open the phoneme abstraction, moving to a learned set of
finer-grained phones.

\newcite{cotterell-eisner:2017:ACL2017} proposed (among other options) using a {\em determinantal point process} (DPP) over a universal inventory $\bigV$ of 53 symbolic (IPA) vowels.  A draw from such a DPP is a language-specific inventory of vowel {\em phonemes}, $V \subseteq \bigV$.  In this paper, we say that a language instead draws its inventory from a larger set $\bigVbar$, again using a DPP.  In both cases, the reason to use a DPP is that it prefers relatively diverse inventories whose individual elements are relatively quantal.

While we could in principle identify $\bigVbar$ with $\mathbb{R}^d$, for convenience we still take it to be a (large) discrete finite set $\bigVbar = \{\vbar_1,\ldots,\vbar_N\}$, whose elements we call {\em phones}. $\bigVbar$ is a learned cross-linguistic parameter of our model; thus, its elements---the ``universal phones''---may or may not correspond to phonetic categories traditionally used by linguists.

We presume that language $\ell$ draws from the DPP a subset $\Vbar\l \subseteq \bigVbar$, whose size we call $n\l$.  For each universal phone $\vbar_i$ that appears in this inventory $\Vbar\l$, the language then draws an observable language-specific pronunciation $\vv\l_i \sim {\cal N}\left(\vmu_i, \sigma^2 I\right)$ from a distribution associated cross-linguistically with the universal phone $\vbar_i$.
We now have an inventory of pronunciations.

As a final step in generating the vowel inventory, we could model IPA labels.  For each $\vbar_i \in \Vbar\l$, a field linguist presumably draws the IPA label $v\l_i$ conditioned on all the pronunciations $\{\vv\l_i \in \mathbb{R}^d: \vbar_i \in \Vbar\l\}$ in the inventory (and perhaps also on their underlying phones $\vbar_i \in\Vbar\l$).  This labeling process may be complex.  While each pronunciation in $\mathbb{R}^d$ (or each underlying phone in $\bigVbar$) may have a preference for certain IPA labels in $\bigV$, the $n\l$ labels must be drawn jointly because the linguist will take care not to use the same label for two phones, and also because the linguist may like to describe the inventory using a small number of distinct IPA features, which will tend to favor factorial grids of symbols.  The linguist's use of IPA features may also be informed by phonological and phonetic processes in the language.  We leave modeling of this step to future work; so our current likelihood term ignores the evidence contributed by the IPA labels in the dataset, considering only the pronunciations in $\mathbb{R}^d$.

The overall idea is that human languages $\ell$ draw their inventories from some universal prior, which we are attempting to reconstruct.  A caveat is that we will train our method by maximum-likelihood, which does not quantify our uncertainty about the reconstructed parameters.  An additional caveat is that some languages in our dataset are related to one another, which belies the idea that they were drawn independently.  Ideally, one ought to capture these relationships using hierarchical or evolutionary modeling techniques.

  \section{Determinantal Point Processes}

\Ryan{There are infelicities in this exposition that I need to work out.
  Specifically, the probability kernel does not inherently produce a
  Gram matrix, whose eigenspectrum lies in $[0, 1]$, so I renormalize. I think
  this makes it wrong to describe it as a point process over an infinite set. It's still a valid geneative model, though.}
Before delving into our generative model, we briefly review technical
background used by \newcite{cotterell-eisner:2017:ACL2017}.
A DPP is a probability distribution over the subsets of a {\em fixed ground set} of size $N$---in our case, the set of phones $\bigVbar$.  The DPP
is usually given as an $L$-ensemble \cite{borodin2005eynard}, meaning that it is parameterized by a positive semi-definite matrix $L \in \mathbb{R}^{N \times N}$. Given
a discrete base set $\bigVbar$ of phones, the probability of a subset $\Vbar \subseteq \bigVbar$ is given by
\begin{equation}
  p(\Vbar) \propto \text{det}\left(L_{\Vbar} \right),
\end{equation}
where $L_{\Vbar}$ is the submatrix of $L$ corresponding to the rows
and columns associated with the subset $\Vbar \subseteq \bigVbar$.
The entry $L_{ij}$, where $i \neq j$, has the effect of describing the similarity between the elements $\vbar_i$ and
$\vbar_j$ (both in $\bigVbar$)---an ingredient needed to
model dispersion. And, the entry $L_{ii}$ describes the quality---focalization---of the vowel $\vbar_i$, i.e., how much the model wants to have $\vbar_i$ in a sampled set independent of the other members.

\begin{figure*}
\centering
\begin{align}\label{eq:likelihood}
&\prod_{\ell=1}^M \Big[ p(\vv^{\ell,1}, \ldots, \vv^{\ell,n\l}  \mid \vmu_1, \ldots, \vmu_N, N)\mathclap{\phantom{\vv\l_1}} \Big]  p(\vmu_1, \ldots \vmu_N \mid N ) \, p(N) \\
&\qquad = \prod_{\ell=1}^M \raisebox{-.1cm}{\Bigg[} \sum_{\va\l \in A(n\l, N)} \Bigg(\prod_{k=1}^{n\l}  \underbrace{p(\vv^{\ell,k}  \mid \vmu_{a\l_k})}_{\text{\numberingBlueB{4}}}\Bigg)  \underbrace{p(\Vbar(\va\l) \mid \vmu_1, \ldots, \vmu_N, N)}_{\numberingGreenB{3}} \raisebox{-.1cm}{\Bigg]} \underbrace{p(\vmu_1, \ldots \vmu_N \mid N)}_{\numberingRedB{2}} \underbrace{p(N)}_{\numberingYellowB{1}} \nonumber
\end{align}
\caption{Joint likelihood of $M$ vowel systems under our deep generative probability model for continuous-space vowel inventories.  Here language $\ell$ has an observed inventory of pronunciations $\{\vv^{\ell,k}: 1 \leq k \leq n\l\}$, and $a\l_k \in [1,N]$ denotes a phone that might be responsible for the pronunciation $\vv^{\ell,k}$.  Thus, $\va\l$ denotes some way to jointly label all $n\l$ pronunciations with distinct phones.  We must sum over all ${N\choose n\l}$ such labelings $\va\l \in A(n^\ell, N)$ since the true labeling is not observed.  In other words, we sum over all ways $\va\l$ of completing the data for language $\ell$.  Within each summand, the product of factors 3 and 4 is the probability of the completed data, i.e., the joint probability of generating the inventory $\Vbar(\va\l)$ of phones used in the labeling and their associated pronunciations.  Factor 3 considers the prior probability of $\Vbar(\va\l)$ under the DPP, and factor 4 is a likelihood term that considers the probability of the associated pronunciations.}
\label{fig:likelihood}
\end{figure*}

\subsection{Probability Kernel}\label{sec:probability-kernel}
In this work, each phone $\vbar_i \in \bigVbar$ is associated with
a probability density over the space of possible pronunciations $\mathbb{R}^2$.
Our measure of phone similarity will consider the ``overlap'' between
the densities associated with two phones. This
works as follows: Given two densities $f(x,y)$
and $f'(x,y)$ over
$\mathbb{R}^2$, we define the kernel \cite{JebaraKH04} as
\begin{equation}\label{eq:probability-kernel}
  {\cal K}(f, f'; \rho) = \int_x \int_y f(x,y)^\rho f'(x,y)^\rho dx\,dy,
\end{equation}
with inverse temperature parameter $\rho$.

In our setting, $f,f'$ will both be Gaussian distributions with means $\vmu$ and $\vmu'$ that share a fixed spherical
covariance matrix $\sigma^2I$.  Then \cref{eq:probability-kernel} and indeed its generalization to any $\mathbb{R}^d$
has a closed-form solution \cite[\S 3.1]{JebaraKH04}:
\begin{align}\label{eq:dispersion}
  {\cal K}(f, & f'; \rho) = \\
  & \left(2\rho\right)^{\frac{d}{2}}\left(2\pi\sigma^2\right)^{\frac{(1 - 2\rho)d}{2}}\exp\left(-\frac{\rho ||\vmu - \vmu'||^2}{4\sigma^2}  \right). \nonumber
\end{align}

\noindent Notice that making $\rho$ small (i.e., high temperature) has an effect on \eqref{eq:dispersion} similar to scaling the variance $\sigma^2$ by the temperature, but it also results in changing the scale of ${\cal K}$, which affects the balance between dispersion and focalization in \eqref{eq:L} below.\jason{This scale factor is a strange $\sigma^2$-dependent function of $\rho$.  Should we parameterize differently, just taking $\rho=1$ but placing an explicit coefficient on $F$ in \eqref{eq:L}?  Would that make it easier to sweep hyperparameters?}

\subsection{Focalization Score}

The probability kernel given in \cref{eq:probability-kernel} naturally handles the linguistic notion of dispersion. What about focalization?  We say that a phone is focal to the extent that it has a high score
\begin{align}\label{eq:focalization}
F(\vmu) &= \exp\left(U_2 \tanh(U_1 \vmu + \mathbf{b}_1) + \mathbf{b}_2 \right) > 0
\end{align}
where $\vmu$ is the mean of its density.  To learn the parameters of this neural network from data is to learn which phones are focal.  We use a neural network since the focal regions of $\mathbb{R}^2$ are distributed in a complex way.

\subsection{The $L$ Matrix}

If $f_i = {\cal N}(\vmu_i,\sigma^2 I)$ is the density associated with the phone $\vbar_i$, we may populate an $N \times N$ real matrix $L$ where
\begin{align}\label{eq:L}
  L_{ij} &= \begin{cases}
    {\cal K}(f_i,f_j; \rho) & \text{if }i \neq j \\
    {\cal K}(f_i,f_j; \rho) + F(\vmu_i) & \text{if }i = j
  \end{cases}
\end{align}

Since $L$ is the sum of two positive definite matrices (the first specializes a known kernel and the second is diagonal and positive), it is also positive definite.  As a result, it can be used to parameterize a DPP over $\bigVbar$.  Indeed, since $L$ is positive definite and not merely positive semidefinite, it will assign positive probability to {\em any} subset of $\bigVbar$.

As previously noted, this DPP does not define a distribution over an infinite set, e.g.,
the powerset of $\mathbb{R}^2$, as does  recent work on continuous DPPs
\cite{affandi2013approximate}. Rather,
it defines a distribution over the powerset of a \emph{set of densities with finite cardinality}.
Once we have sampled a subset of densities, a real-valued quantity may be additionally
sampled from each sampled density.

\begin{algorithm}[t]
  \begin{algorithmic}[1]
\State $N \sim \text{Poisson}\left(\lambda\right)$ $(\in \mathbb{N})$ \hfill \numberingYellow{1}
\For{$i=1$ {\bf to} $N$}
\State $\vmu_i \sim {\cal N}\left(\mathbf{0}, I\right)$ $(\in \mathbb{R}^2)$ \hfill \numberingRed{2}
\EndFor
\State define $L \in \mathbb{R}^{N\times N}$ via \eqref{eq:L}
\For{$\ell=1$ {\bf to} $M$}
\State  $\Vbar\l \sim \text{DPP}\left(L\right)$ $(\subseteq [1,N])$; let $n\l=|\Vbar\l|$ \hfill \numberingGreen{3}
\For{$i \in \Vbar\l$}
\State $\vvtilde\l_i \sim {\cal N}\left(\vmu_i, \sigma^2 I\right)$ \hfill \numberingBlue{4}
\State $\vv\l_i = \NN\left(\vvtilde_i\l\right)$ \hfill \numberingBlue{4}
\EndFor
\EndFor
\end{algorithmic}
  \caption{Generative Process}
  \label{alg:generative-process}
\end{algorithm}

\section{A Deep Generative Model}\label{sec:model}

We are now in a position to expound our generative model of
continuous-space vowel typology. We generate a set of formant pairs
for $M$ languages in a four step process. Note
that throughout this exposition, language-specific quantities
with be superscripted with an integral language marker $\ell$, whereas
universal quantities are left unsuperscripted.
The generative process is written in algorithmic
form in \cref{alg:generative-process}.
Note
that each step is numbered and color-coded for ease
of comparison with the full joint likelihood in \cref{fig:likelihood}.

\paragraph{Step \numberingYellow{1}: $p(N)$.}
We sample the size $N$ of the universal phone inventory $\bigVbar$
from a Poisson distribution with a rate parameter $\lambda$, i.e.,
\begin{equation}
  N \sim \text{Poisson}\left(\lambda\right).
\end{equation}
That is, we do not presuppose a certain number of phones in the model.

\paragraph{Step \numberingRed{2}: $p(\vmu_1, \ldots, \vmu_N)$.}
Next, we sample the means $\vmu_i$ of the Gaussian phones.
In the model presented here, we assume that each phone is generated
independently, so $p(\vmu_1, \ldots, \vmu_N) = \prod_{i=1}^N p(\vmu_i)$.
Also, we assume a standard Gaussian prior
over the means,
$
  \vmu_i \sim {\cal N}(\mathbf{0}, I).
$

The sampled means define our $N$ Gaussian phones
${\cal N}\left(\vmu_i, \sigma^2 I\right)$: we are assuming
for simplicity that all phones share a {\em single} spherical
covariance matrix, defined by the hyperparameter $\sigma^2$.  The
dispersion and focalization of these phones define the matrix $L$
according to \crefrange{eq:dispersion}{eq:L}, where $\rho$ in
\eqref{eq:dispersion} and the weights of the focalization neural net
\eqref{eq:focalization} are also hyperparameters.

\paragraph{Step \numberingGreen{3}: $p(\Vbar\l  \mid \vmu_1, \ldots, \vmu_N)$.}

Next, for each language $\ell \in [1, \ldots, M]$, we sample a diverse subset of the $N$ phones,
via a single draw from a DPP parameterized by matrix $L$:
\begin{equation}
\Vbar\l \sim \text{DPP}(L),
\end{equation}
where $\Vbar\l \subseteq [1,N]$.  Thus, $i \in \Vbar\l$ means that language $\ell$ contains phone $\vbar_i$.
Note that even the size of the inventory, $n\l = |\Vbar\l|$, was chosen by the DPP.
In general, we have $n\l \ll N$.

\paragraph{Step \numberingBlue{4}: $\prod_{i\in\Vbar\l} p(\vv\l_i  \mid \vmu_i)$}
The final step in our generative process is that
the phones $\vbar_i$ in language $\ell$ must generate the
pronunciations $\vv\l_i \in \mathbb{R}^2$ (formant vectors) that are actually observed in language $\ell$.
Each vector takes two steps.  For each $i \in \Vbar\l$, we generate
an underlying $\vvtilde_i \in \mathbb{R}^2$ from the corresponding Gaussian phone.
Then, we run this vector through a feed-forward neural network $\NN$ with parameters
$\vtheta$.  In short:
\begin{align}
  \vvtilde\l_i & \sim {\cal N}(\vmu_i, \sigma^2 I) \\
  \vv\l_i &= \NN(\vvtilde\l_i), \label{eq:neural}
\end{align}
where the second step is deterministic.  We can fuse these two steps into
a single step $p(\vv_i \mid \vmu_i)$, whose closed-form density is
given in \cref{eq:neural-final} below.  In effect, step 4 takes a Gaussian
phone as input and produces the observed formant vector with an
{\em underlying} formant vector in the middle.

This completes our generative process.  We do not observe all the steps, but only the final collection of pronunciations $\vv\l_i$ for each language, where the subscripts $i$ that indicate phone identity have been lost.  The probability of this incomplete dataset involves summing over possible phones for each pronunciation, and is presented in \cref{fig:likelihood}.

\subsection{A Neural Transformation of a Gaussian}\label{sec:transformation}
A crucial bit of our model is running a sample from a Gaussian through
a neural network. Under certain restrictions, we can find a closed
form for the resulting density; we discuss these below.
Let $\NN$ be a depth-2 multi-layer
perceptron
\begin{equation}
 \NN(\mathbf{\vvtilde}_i) = W_2 \tanh \left( W_1 \mathbf{\vvtilde}_i + \mathbf{b}_1 \right) + \mathbf{b}_2.
\end{equation}
In order to find a closed-form solution, we require that \eqref{eq:focalization}
be a diffeomorphism, i.e., an invertible mapping from $\mathbb{R}^2
\to \mathbb{R}^2$ where both $\NN$ and its inverse $\NN^{-1}$ are
differentiable. This will be true as long as $W_1, W_2 \in \mathbb{R}^{2
  \times 2}$ are square matrices of full-rank and we choose a smooth,
invertible activation function, such as $\tanh$.  Under those
conditions, we may apply the standard theorem for transforming a
random variable \citep[see][]{leon2008probability}:
\begin{align}
  p(\vv_i \mid \vmu_i)
  &= p(\NN^{-1}(\vv_i) \mid \vmu_i)\,\, \text{det}\, J_{\NN^{-1}(\vv_i)} \nonumber \\
  &= p(\vvtilde_i \mid \vmu_i)\,\, \text{det}\, J_{\NN^{-1}(\vv_i)} \label{eq:neural-final}
\end{align}
where $J_{\NN^{-1}(\xx)}$ is the Jacobian of the inverse of the neural network at the point $\xx$.
Recall that $p(\vvtilde_i \mid \vmu_i)$ is Gaussian-distributed.

\section{Modeling Assumptions}\label{sec:modeling-assumptions}
Imbued in our generative story are a number
of assumptions about the linguistic processes behind vowel inventories.
We briefly draw connections between our theory and the linguistics literature.

\paragraph{Why underlying phones?}
A technical assumption of our model is the existence of a universal
set of underlying phones.  Each phone is equipped with a probability
distribution over reported acoustic measurements (pronunciations), to
allow for a single phone to account for multiple slightly different
pronunciations in different languages (though never in the same
language).  This distribution can capture both actual interlingual
variation and also random noise in the measurement process.

While our universal phones may seem to resemble the universal IPA
symbols used in phonological transcription, they lack the rich
featural specifications of such phonemes.  A phone in our model has no
features other than its mean position, which wholly determines its
behavior.  Our universal phones are not a substantive linguistic
hypothesis, but are essentially just a way of partitioning
$\mathbb{R}^2$ into finitely many small regions whose similarity and
focalization can be precomputed.  This technical trick allows us to
use a discrete rather than a continuous DPP over the
$\mathbb{R}^2$ space.\footnote{\label{fn:overfit}Indeed, we could have
  simply taken our universal phone set to be a huge set of tiny,
  regularly spaced overlapping Gaussians that ``covered'' (say) the
  unit circle.  As a computational matter, we instead opted to use a
  smaller set of Gaussians, giving the learner the freedom to infer
  their positions and tune their variance $\sigma^2$.  Because of this
  freedom, this set should not be too large, or a MAP learner may
  overfit the training data with zero-variance Gaussians and be unable
  to explain the test languages---similar to overfitting a Gaussian
  mixture model.}

\paragraph{Why a neural network?}
Our phones are Gaussians of spherical variance $\sigma^2$, presumed to
be scattered with variance 1 about a two-dimensional {\em latent}
vowel space.  Distances in this latent space are used to compute the
dissimilarity of phones for modeling dispersion, and also to describe
the phone's ability to vary across languages.  That is, two phones
that are {\em distant} in the latent space can appear in the same
inventory---presumably they are easy to discriminate in both
perception and articulation---and it is easy to choose which one
better explains an acoustic measurement, thereby affecting the
other measurements that may appear in the inventory.

We relate this {\em latent} space to measurable acoustic space by a
learned diffeomorphism $\NN$ \cite{cotterell-eisner:2017:ACL2017}.
$\NN^{-1}$ can be regarded as warping the acoustic distances into
perceptual/articulatory distances.  In some ``high-resolution''
regions of acoustic space, phones with fairly similar $(F_1,F_2)$
values might yet be far apart in the latent space.  Conversely, in
other regions, relatively large acoustic changes in some direction
might not prevent two phones from acting as similar or two
pronunciations from being attributed to the same phone.  In general, a
unit circle of radius $\sigma$ in latent space may be mapped by $\NN$
to an oddly shaped connected region in acoustic space, and a Gaussian
in latent space may be mapped to a multimodal distribution.

\section{Inference and Learning}\label{sec:inference-learning}
We fit our model via MAP-EM \cite{dempster1977maximum}. The E-step involves deciding which phones each language has. To achieve this, we fashion a Gibbs sampler \cite{geman1984stochastic}, yielding a Markov-Chain
Monte Carlo E-step \cite{levine2001implementations}.

\subsection{Inference: MCMC E-Step}

Inference in our model is intractable even when the phones $\vmu_1, \ldots,
\vmu_N$ are fixed.  Given a language with $n$
vowels, we have to determine which subset of the $N$ phones
best explains those vowels. As discussed above, the alignment $\va$
between the $n$ vowels and $n$ of the $N$ phones represents a
latent variable. Marginalizing it out is \#P-hard, as we can see that it
is equivalent to summing over all bipartite matchings in a weighted
graph, which, in turn, is as costly as computing
the permanent of a matrix \cite{valiant1979complexity}. Our sampler\footnote{Taken from \newcite[3.1]{volkovs2012efficient}.}
is an approximation
algorithm for the task. We are interested in sampling
 $\va$, the labeling of observed vowels
with universal phones.  Note that this implicitly
samples the language's phone inventory $\Vbar(\va)$, which is fully determined by $\va$.

Specifically, we employ an MCMC method closely related to Gibbs sampling.  At each step of the sampler,
we update our vowel-phone alignment $\va\l$ as follows.  Choose a language $\ell$ and a vowel index
$k \in [1,n\l]$, and let $i = a\l_k$ (that is, pronunciation
$\vv^{\ell,k}$ is currently labeled with universal phone $\vbar_i$).
We will consider changing $a\l_k$ to $j$, where $j$ is drawn
from the $(N - n\l)$ phones that do {\em not} appear in
$\Vbar(\va\l)$, heuristically
choosing $j$ in proportion to the likelihood $p(\vv^{\ell,k} \mid
\vmu_{j})$.
We then stochastically decide whether to keep $a\l_k = i$ or set $a\l_k = j$ in
proportion to the resulting values of the product $\text{\numberingBlueB{4}}\cdot\text{\numberingGreenB{3}}$ in
\cref{eq:likelihood}.

For a single E-step, the Gibbs sampler ``warm-starts'' with the labeling from the end of the previous iteration's E-step.  It sweeps $S=5$ times through all vowels for all languages, and returns $S$ sampled labelings, one from the end of each sweep.

We are also interested in automatically choosing the number of phones
$N$, for which we take the Poisson's rate parameter $\lambda = 100$.
To this end, we employ reversible-jump MCMC
\cite{green1995reversible}, resampling $N$ at the start of every
E-step. \jason{Is this right?  and how many split-merge moves do you
  allow when resampling $N$?  you said originally that you sampled $N$
  once per ``iteration,'' but it wasn't clear whether ``iteration''
  meant that you did it once per E step or $S$ times per E step.  I am
  guessing the former, since think the M-step is incoherent if you
  have split and merged phones during the E step.}

\subsection{Learning: M-Step}
Given the set of sampled alignments provided by the E-step, our M-step
consists of optimizing the log-likelihood of the now-complete training data using
the inferred latent variables. \jason[color=lightgray]{since $S$ is small, perhaps should have used
an online EM method (Percy Liang) that also includes some samples from previous E-steps}
We achieved this through SGD training of the diffeomorphism parameters $\vtheta$,
the means $\vmu_i$ of the Gaussian phones, and the parameters of the
focalization kernel ${\cal F}$. \jason{any regularization???}

\section{Experiments}

\subsection{Data}
Our data is taken from the Becker-Kristal corpus
\cite{becker2006predicting}, which is a compilation of various
phonetic studies and forms the largest multi-lingual phonetic
database.
Each entry in the corpus corresponds to a linguist's phonetic description
of a language's vowel system: an inventory consisting of IPA symbols
where each symbol is associated with two or more formant values.  The
corpus contains data from 233 distinct languages.  When multiple
inventories were available for the same language (due to various
studies in the literature), we selected one at random and
discarded the others.\jason[color=yellow]{how many training, dev, and test languages??}

\subsection{Baselines}

\paragraph{Baseline \#1: Removing dispersion.}

The key technical innovation in our work lies in the incorporation of
a DPP into a generative model of vowel formants---a continuous-valued
quantity. The role of the DPP was to model the linguistic principle of
dispersion---we may cripple this portion of our model, e.g., by
forcing ${\cal K}$ to be a diagonal kernel, i.e., $K_{ij} = 0$ for
$i \neq j$.  In this case the DPP becomes a Bernoulli Point Process
(BPP)---a special case of the DPP. Since dispersion is widely accepted
to be an important principle governing naturally occurring vowel
systems, we expect a system trained without such knowledge to
perform worse.

\paragraph{Baseline \#2: Removing the neural network $\NN$.}
Another question we may ask of our formulation is whether we actually
need a fancy neural mapping $\NN$ to model our typological data
well. The human perceptual system is known to perform a non-linear
transformation on acoustic signals, starting with the non-linear
cochlear transform that is physically performed in the ear.
While $\NN^{-1}$ is intended as loosely analogous, we determine
its benefit by removing \cref{eq:neural} from our generative story, i.e.,
we take the observed formants $\vv_k$ to arise directly from the
Gaussian phones.

\paragraph{Baseline \#3: Supervised phones and alignments.}
A final baseline we consider is {\em supervised} phones.  Linguists
standardly employ a finite set of phones---symbols from the
international phonetic alphabet (IPA). In phonetic annotation, it
is common to map each sound in a language back to this universal
discrete alphabet. Under such an annotation scheme, it is easy to
discern, cross-linguistically, which vowels originate from the same
phoneme: an /\textipa{I}/ in German may be roughly equated with an
/\textipa{I}/ in English. However, it is not clear how consistent this
annotation truly is. There are several reasons to expect high-variance
in the cross-linguistic acoustic signal. First, IPA symbols are
primarily useful for interlinked phonological distinctions, i.e., one
applies the symbol /\textipa{I}/ to distinguish it from /i/ in the
given language, rather than to associate it with the sound bearing the
same symbol in a second language. Second, field linguists often resort
to the closest common IPA symbol, rather than an exact match: if a
language makes no distinction between /i/ and /\textipa{I}/, it is
more common to denote the sound with a /i/. Thus, IPA may not be as
universal as hoped. Our dataset contains 50 IPA symbols so this
baseline is only reported for $N=50$.

\begin{table}
  \centering
  \begin{adjustbox}{width=1.\columnwidth}
  \begin{tabular}{ll llll} \toprule
    $N$ & metric & DPP$+\NN$  & BPP$+\NN$ & DPP$-\NN$  & Sup.  \\ \midrule
        & x-ent    & 540.02 & 540.05 & 600.34     &  \xmark  \\
    $15$  & cloze1 & 5.76 & 5.76 & 6.53 & \xmark \\
	& cloze12 & 4.89 & 4.89 & 5.24 & \xmark \\ \midrule
    & x-ent & 280.47 & 275.36 & 335.36 & \xmark  \\
    $25$  & cloze1 &  5.04 & 5.25 & 6.23 & \xmark \\
	& cloze12 & 4.76 & 4.97 & 5.43 & \xmark \\ \midrule
    & x-ent & 222.85 & 231.70 & 320.05 &  1610.37\\
      $50$ & cloze1 & 3.38 & 3.16 & 4.02 & 4.96  \\
	& cloze12 & 2.73 & 2.93 & 3.04 &  6.95 \\ \midrule
    & x-ent & 212.14 & 220.42  & 380.31 & \xmark\\
          $57$ & cloze1 & 2.21 & 3.08 & 3.25 & \xmark  \\
	& cloze12 & 2.01 & 3.05 & 3.41 &  \xmark  \\ \midrule
    & x-ent & 271.95 & 301.45  & 380.02 & \xmark\\
    $100$  &  cloze1 & 2.26 & 2.42 & 3.03 & \xmark\\
 	& cloze12 & 1.96 & 2.01 & 2.51 & \xmark \\ \midrule
    \bottomrule
  \end{tabular}
  \end{adjustbox}
  \caption{Cross-entropy in nats per language (lower is better) and expected
    Euclidean-distance error of
    the cloze prediction (lower is better).  The overall best value for each task
    is boldfaced.  The case $N=50$ is compared against our supervised baseline.
    The $N = 57$ row is the case where we allowed $N$ to fluctuate during inference using reversible-jump
    MCMC; this was the $N$ value selected at the final EM iteration.}
  \label{tab:results}
  \Jason[color=yellow]{rerun these results with harmonic mean and the
    other changes, expecially those in yellow.  Don't forget to change the number 53.
  To speed up the experiments, consider dropping $N=1000$ and changing
  $N=500$ to $N=200$.}
\Jason[color=yellow]{significance tests??}
\end{table}

\subsection{Evaluation}
Evaluation in our setting is tricky. The scientific
goal of our work is to place a bit of linguistic theory on
a firm probabilistic footing, rather than a downstream
engineering-task, whose performance we could measure.
We consider three metrics.

\paragraph{Cross-Entropy.}
Our first evaluation metric is cross-entropy: the average negative
log-probability of the vowel systems in held-out test data, given the
universal inventory of $N$ phones that we trained through EM.  We find
this to be the cleanest method for scientific evaluation---it is the
metric of optimization and has a clear interpretation: how surprised
was the model to see the vowel systems of held-out, but attested,
languages?

The cross-entropy is the negative log of the
$\prod \big[ \cdots \big]$ expression in \cref{eq:likelihood}, with
$\ell$ now ranging over held-out languages.\footnote{Since that
  expression is the product of both probability distributions and
  probability densities, our ``cross-entropy'' metric is actually the sum of
  both entropy terms and (potentially negative) differential entropy
  terms.  Thus, a value of 0 has no special significance.}
\newcite{wallach_evaluation_2009} give several methods for estimating
the intractable sum in language $\ell$.  We use the simple harmonic mean estimator, based
on 50 samples of $\va\l$ drawn with our Gibbs sampler (warm-started from the
final E-step of training).

\paragraph{Cloze Evaluation.}
In addition, following \newcite{cotterell-eisner:2017:ACL2017}, we evaluate
our trained model's ability to perform a cloze task
\cite{taylor1953cloze}.  Given $n\l-1$ or $n\l-2$ of the vowels in
held-out language $\ell$, can we predict the pronunciations $\vv_k$ of
the remaining 1 or 2?  We predict $\vv_k$ to be $\NN(\vmu_i)$ where
$i = a\l_k$ is the phone inferred by the sampler.  Note that the sampler's
inference here is based only on the observed vowels (the likelihood) and
the focalization-dispersion preferences of the DPP (the prior).
We report the expected error of such a prediction---where error
is quantified by Euclidean distance in $(F_1,F_2)$ formant
space\jason{not squared Euclidean distance, which would also be
  common?}---over the same 50 samples of $\va\l$.

For instance, consider
a previously unseen vowel system with formant values
$\{$$(499,2199)$,
$(861,1420)$,
$(571,1079)$$\}$.
A ``cloze1'' evaluation would aim to predict $\{(499,2199)\}$ as the missing
vowel, given $\{$$(861,1420)$,
$(571,1079)$$\}$, and the fact that $n\l=3$.  A ``cloze12'' evaluation
would aim to predict two missing vowels.

\begin{figure}
  \begin{adjustbox}{width=\columnwidth}
    \includegraphics{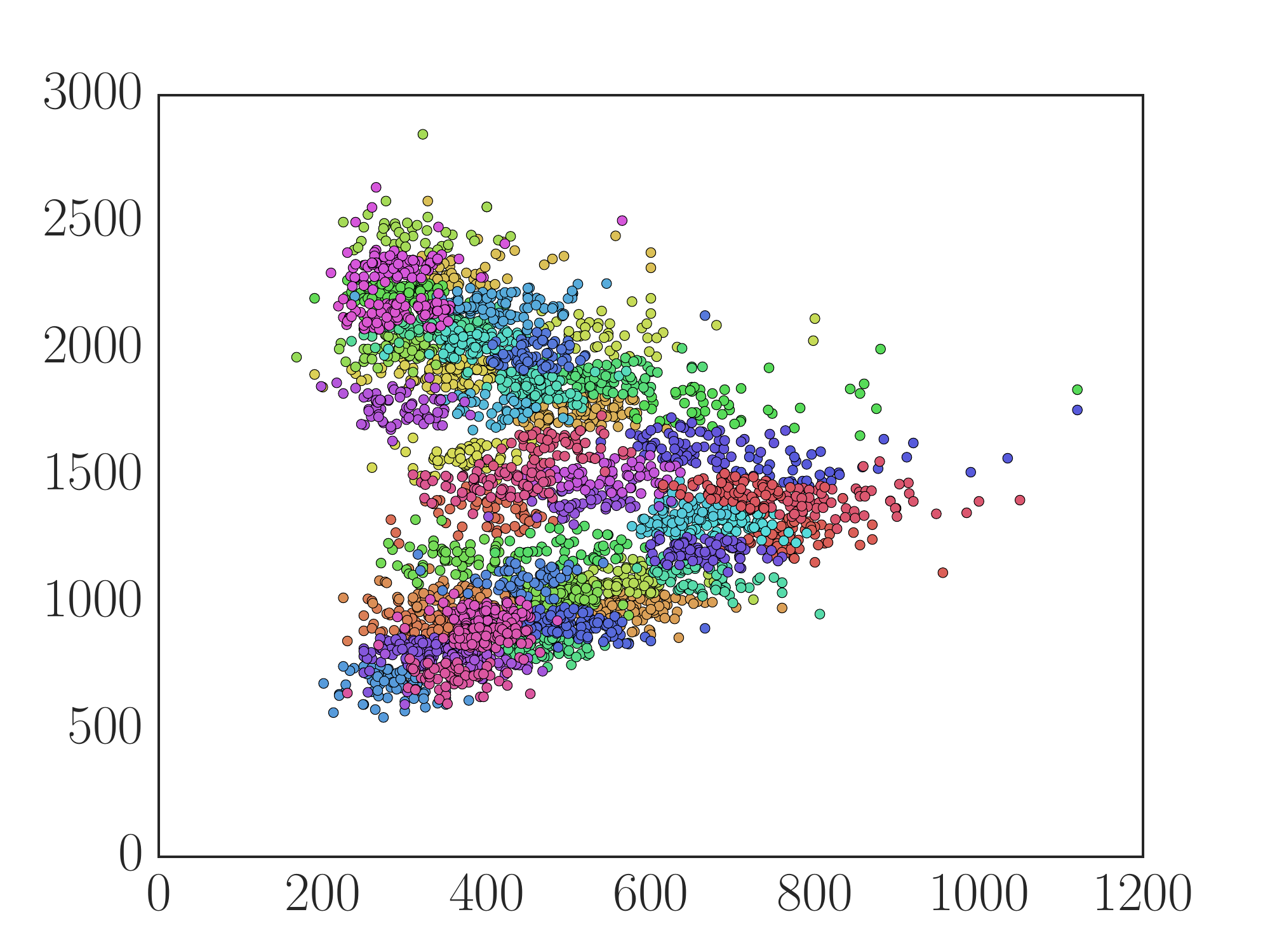}
  \end{adjustbox}
  \caption{A graph of $\vv = (F_1, F_2)$ in the union of all the
    training languages' inventories, color-coded by inferred phone ($N = 50$).
   }
   \Jason[color=yellow]{update the graph from the new experiments!
     And say something like ``While it may be hard to perceive the
     fine-grained color differences, 43 different colors are used in
     this graph; phones with nearby means have been assigned
     contrasting colors.''  I added the word ``inferred'': this is
     {\em inferred} phone and not {\em supervised} phone, right?}
  \label{fig:graph}
\end{figure}

\subsection{Experimental Details}
Here, we report experimental details and the hyperparameters
that we use to achieve the results reported. We consider
a neural network $\NN$ with $k \in [1, 4]$ layers  and find $k=1$
the best performer on development data. Recall
that our \emph{diffeomorphism} constraint requires that each layer
have exactly two hidden units, the same as the number of observed formants. We consider $N \in \{15, 25, 50, 100\}$ phones as well as letting $N$ fluctuate with
reversible-jump MCMC (see \cref{fn:overfit}).
We train for 100 iterations of EM, taking $S=5$ samples at each
E-step.
At each M-step, we run 50 iterations of SGD for the focalization NN
and also for the diffeomorphism NN.
For each $N$, we selected $(\sigma^2,\rho)$ by minimizing
cross-entropy on a held-out development set. We considered
$(\sigma^2, \rho) \in \{10^k \}_{k=1}^5 \times \{\rho^k\}_{k=1}^5$. \jason[color=yellow]{make
sure to do this, and report the values you swept}

\subsection{Results and Error Analysis}
We report results in \cref{tab:results}. We find that our DPP
model improves over the baselines. The results support two claims: (i) dispersion
plays an important role in the structure of vowel systems and (ii) learning a non-linear transformation of a Gaussian improves our ability to model sets of formant-pairs. Also, we observe that as we increase the number of phones, the role
of the DPP becomes more important.
We visualize a sample of the trained alignment in
\cref{fig:graph}.
\jason[color=yellow]{update the discussion below after the new experiments}

\paragraph{Frequency Encodes Dispersion.}
Why does dispersion not always help? The models
with fewer phones do not reap the benefits
that the models with more phones do. The reason lies
in the fact that the most common vowel formants are \emph{already} dispersed. This indicates that we
still have not quite modeled the mechanisms that select
for good vowel formants, despite our work at the phonetic level;
further research is needed. We would prefer a model
that explains the \emph{evolutionary motivation} of sound systems
as communication systems.

\paragraph{Number of Induced Phones.}
What is most salient in the number of induced phones
is that it is close to the number of IPA phonemes
in the data. However,
the performance of the phoneme-supervised system is much worse,
indicating that, perhaps, while the linguists have the right
idea about the \emph{number} of universal symbols, they did not specify the correct IPA symbol in all cases. Our data analysis indicates
that this is often due to pragmatic concerns in linguistic field analysis. For example, even if /\textipa{I}/ is the proper IPA symbol for the sound, if there is no other sound in the vicinity
the annotator may prefer to use more common /i/.

\section{Related Work}
Most closely related to our work is the classic study of
\newcite{liljencrants1972numerical}, who provide a simulation-based
account of vowel systems.
They argued that minima of a certain objective that encodes dispersion
should correspond to canonical vowel systems of a given size $n$. Our
tack is different in that we construct a generative probability model,
whose parameters we learn from data. However, the essence of modeling
is the same in that we explain \emph{formant} values, rather than
discrete IPA symbols. By extension, our work is also closely related
to extensions of this theory
\cite{schwartz1997dispersion,roark2001explaining} that focused on
incorporating the notion of focalization into the experiments.

Our present paper can also be regarded as a continuation of
\newcite{cotterell-eisner:2017:ACL2017}, in which we used DPPs to
model vowel inventories as sets of discrete IPA symbols.  That paper
pretended that each IPA symbol had a single cross-linguistic
$(F_1,F_2)$ pair, an idealization that we remove in this paper by
discarding the IPA symbols and modeling formant values directly.

\section{Conclusion}

Our model combines existing techniques of probabilistic modeling and
inference to attempt to fit the actual distribution of the world's
vowel systems.  We presented a generative probability model of sets of
measured $(F_1,F_2)$ pairs.  We view this as a necessary step in the
development of generative probability models that can explain the
distribution of the world's languages.  Previous work on generating
vowel inventories has focused on how those inventories were
transcribed into IPA by field linguists, whereas we focus on the field
linguists' acoustic measurements of how the vowels are actually
pronounced.

\section*{Acknowledgments}
We would like to acknowledge Tim Vieira, Katharina Kann, Sebastian
Mielke and Chu-Cheng Lin for reading many early drafts. The first
author would like to acknowledge an NDSEG grant and a Facebook PhD
fellowship.  This material is also based upon work supported by the National
Science Foundation under Grant No.\@ 1718846 to the last author.

\bibliography{continuous-dpp}
\bibliographystyle{acl_natbib}

\end{document}